\documentstyle[pstricks,pst-node]{llncs}

\begin{document}

\title{Robust Processing of Natural Language}
\author{Wolfgang Menzel}
\institute{Fachbereich Informatik, Universit\"at Hamburg \\
Vogt-K\"olln-Stra{\ss}e 30, 22527 Hamburg, Germany}

\maketitle

\begin{abstract}
Previous approaches to robustness in natural language processing
usually treat deviant input by relaxing grammatical
constraints whenever a successful analysis cannot be provided by
``normal'' means. This schema implies, that error detection
always comes prior to error handling, a behaviour which hardly can
compete with its human model, where many erroneous
situations are treated without even noticing them.

The paper analyses the necessary preconditions for achieving a
higher degree of robustness in natural language processing and
suggests a quite different approach based on a
procedure for structural disambiguation. It not only offers the
possibility to cope with robustness issues in a more natural way but
eventually might be suited to accommodate quite different aspects of
robust behaviour within a single framework.

\end{abstract}

\section{Robustness in Natural Language Processing}

The notion of robustness in natural language processing is a rather broad one
and lacks a precise definition. Usually, it is taken to describe a kind of
monotonic behaviour, which should be guaranteed whenever a system is exposed to
some sort of non-standard input data: A comparatively small deviation from a
predefined ideal should lead to no or only minor disturbances in the system's
response, whereas a total failure might only be accepted for sufficiently
distorted input.

Under this informal notion robustness may well be interpreted as a system's
indifference to a wide range of external disruptive factors including

\begin{itemize}
  \item the inherent uncertainty of real world input, e.g. speech or hand
writing,
  \item noisy environments,
  \item the variance between speakers, for instance idiolectal, dialectal or
sociolectal,
  \item ``erroneous'' input with respect to some normative standard,
  \item an insufficient competence of the processing system, if e.g. exposed to
a non-native language or new terminology,
  \item highly varying speech rates and
  \item resource limitations due to the parallel execution of several mental
activities.
\end{itemize}

One of the most impressive features of human language processing is the ability
to retain its basic capabilities even if it is exposed to a combination of
adverse
factors. Technical solutions, on the other hand, are likely to have serious
problems if confronted with only a single type of distortion, apart from the
fundamental
difficulties to supply the desired monotonic behaviour at all.

Accordingly, problems of robustness in NLP have almost never been considered
from a unifying perspective so far. A number of very specific techniques for
some of those different aspects has been developed, which hardly can be related
to each other.

Robustness, for instance, is a key issue in speech recognition, where
reliable recognition results for a variety of speakers and speaking
conditions are desired. Two basic technologies attempt to support this goal
\vspace{-2pt}
\begin{itemize}
  \item robust stochastic modelling techniques which are able to capture
generalizations across the individual variety\footnote{The difficulties with a
straightforward generalization of this approach to e.g. syntactic or semantic
anomalies are obvious: It would require huge amounts of sufficiently deviant
utterances being available as training data. This renders the approach
technically infeasible and cognitively implausible. \\ For similar reasons
connectionist approaches are not considered here: At the moment they seem to be
limited to approximate solutions for flat representations (cf.
\cite{wermter:weber:94a}).} and
  \item sophisticated search procedures which select among huge amounts of
competing recognition hypotheses by comparing probability estimations for
signal
segments of increasing length. \vspace{-2pt}
\end{itemize}
Special signal enhancement techniques are used to suppress stationary
environmental noise. There are other aspects of robustness which even have not
been treated at all, including the flexible adaptation to external time
constraints or internal resource limitations.

Traditionally the notion of robustness has been strongly connected to the
processing of ill-formed input\footnote{For a good overview see
\cite{stede:92a}.}, where ill-formedness can be defined both, in terms of human
standards of grammaticality or in terms of unexpected input. Most of the work
has been concerned with the problem from a {\em purely syntactic point of view}
and
usually relied on two basic techniques: error anticipation and constraint
relaxation.

{\em Error anticipation} identifies a number of common mistakes and tries to
integrate
them into the existing grammar by devising dedicated extensions to its
coverage.
Therefore, the method is limited to a few selected types of deviant
constructions which are notorious and therefore predictable, namely
\vspace{-2pt}
\begin{itemize}
  \item stereotypical spelling mistakes ({\em *comittee, *rigth}, etc.),
  \item performance phenomena in spoken language, like restarts (cf.
\cite{goeser:92}) and
  \item interference-based competence errors in early phases of second language
learning (cf. \cite{catt:88}). \vspace{-2pt}
\end{itemize}
Obviously, the complete ``innovative'' potential and the individual creativity
for producing ill-formed input cannot be adequately captured by such means
alone.

On the other hand, {\em constraint relaxation techniques} rely on a systematic
variation of existing grammar rules written for standard input. Initially, the
idea was restricted to the stepwise retraction of e.g. agreement conditions in
syntactic rules. It can easily be extended to incorporate arbitrary rule
transformations in order to allow for the insertion, deletion, substitution and
transposition of elements. The difference vanishes completely within modern
constraint-based formalisms \cite{uszkoreit:91} \cite{erbach:93}, where a
transposition of constituents can be interpreted equally well as a relaxation
of
linear precedence constraints. Furthermore, constraints can be annotated by
their
degree of vulnerability, hence allowing to include aspects of error
anticipation
into the relaxation framework.

Since both, error anticipation and constraint relaxation considerably enlarge
the generative capacity of the original grammar they will lead
to spurious ambiguities and serious search problems. This restricts their
application to a kind of {\em post mortem} analysis\footnote{There are
exceptions to
every rule: For language learning purposes \cite{kudo:etal:88} propose an
initial analysis based on a moderately weak grammar and followed by a more
rigid second pass.}. Only if a failure of the standard analysis procedure
indicates the presence of non-standard input, error rules or relaxation
techniques are activated to integrate the fragmentary results obtained so far.

Even a superficial comparison with human processing principles shows the
fundamental deficit of these approaches. A human reader or listener accepts
ill-formed input to a wide degree, often without noticing an error at all. This
is particularly true if strong expectations concerning the content of the
utterance are involved or if heavy time constraints restrict the processing
depth.

Obviously, there is a fundamental parallelism between robustness issues and
time
considerations, which syntactically oriented solutions lack so far. Robustness
in human language processing does not amount to an additional effort, but
instead facilitates both, insensitivity to ill-formed input as well as a
flexible adaptation to temporal restrictions.

This basic pattern is much better modelled by {\em semantically oriented
approaches}
based on the slot-and-filler-principle. Here, highly domain specific
expectations are coded by means of frame-like structures and checked against
the
input for satisfaction. The schema can be successfully extended to a kind of
skimming understanding bringing together the question of robustness against
syntactically ill-formed input and some simple considerations concerning
resource limitations.

This advantage of a semantically guided analysis, however, is won by the cost
of
excluding another important robustness feature, namely the ability to cope with
unexpected input (e.g. a change of topic beyond the narrow limitations of the
domain or the violation of selectional restrictions in metaphorical
expressions).

\section{Observations from Human Language Processing}

Psycholinguistic evidence provides a contradictory picture of human language
processing. Some observations clearly support a rather strong modular
organization with processing units of great autonomy like syntax and semantics
\cite{forster:79} \cite{frazier:87}. On the other hand there is a
considerable semantic influence on the assignment of syntactic structure
\cite{marslen-wilson:tyler:87} which suggests a highly integrated processing
architecture.

Robust behaviour in natural language understanding seems to require both,
\begin{itemize}
  \item the autonomy between parallel lines of processing which embodies
redundancy
and allows to compensate partial insufficiencies and
  \item the interactive nature of informational exchange which allows to relate
partial structures on different levels of granularity.
\end{itemize}
{\em Functional autonomy} undoubtedly is of fundamental importance for
robustness.
It allows to yield an at least vague interpretation even in cases of extremely
distorted input:
\begin{enumerate}
  \item A semantically almost empty sentence can be analysed quite well by
syntactic means alone, delivering a hypothetical interpretation in terms of a
possible world with highly underspecified referential object descriptions and
possibly ambiguous thematic roles. \\[3pt]
\hspace*{1cm}{\em ``... und grausig gutzt der Golz.''}\cite{morgenstern:33}
\hfill (1) \\[-3pt]
  \item Syntactically ill-formed utterances are interpreted based on semantic
and background knowledge even if subcategorization regularities or other
grammatical constraints are violated.
\end{enumerate}
Although both processing units are -- at least partially -- able to generate
some useful interpretation independently of the other one, best results, of
course, are to be expected if they combine their efforts in a systematic way.

Parallel and autonomous structures in language processing have not only evolved
between syntactic and semantic aspects of language. They can be
observed equally well at the level of speech comprehension where auditory
(hearing) and
visual (lip-reading) clues are usually combined to achieve a reliable
recognition result. Again, both systems -- in principle -- are able to work
independently, but synergy occurs if both are activated concurrently.

A second group of observations related to the question of robustness concerns
the {\em expectation-driven} nature of human language understanding. Here,
expectations come to play at two different dimensions:

\begin{itemize}
  \item Syntactic, semantic and pragmatic predictions about future input
derived
from previous parts of the utterance or dialogue.
  \item Expectations exchanged between parallel and autonomous processing
structures for syntax and semantics.
\end{itemize}
The role of dynamic expectations has mostly been investigated from the
viewpoint of a possible search space reduction in prediction based parsing
strategies
(namely left or head corner algorithms). If used to select between competing
hypotheses in speech recognition the predictive capacity of a grammar
can contribute additionally to an enhanced robustness of the overall system
\cite{hauenstein:weber:94} \cite{hanrieder:goerz:95}.

Although the importance of predictions for robustness is beyond question, here
the second type of expectations shall be examined as a matter of priority,
since they are expected to establish the attempted informational
coupling between parallel processing units. As the simple examples above
have shown, no predefined direction for this exchange of information can be
assumed. Certain syntactic constructions may trigger specific semantic
interpretations, a view which is strongly supported by the traditional
perspective on the
relation between syntax and semantics. In the opposite direction, semantic
relations, e.g. derived from background knowledge, can not only be used to
disambiguate between preestablished syntactic readings, but moreover are able
to
actively propose suitable syntactic structures. This bidirectionality of
interaction seems to be of great importance for the ability to provide the
mutual compensation necessary to treat deviant constructions of different kind.

Of course, the expectation-based nature of natural language processing cannot
guarantee a failure-proof performance under all circumstances\footnote{Note
that
perfect performance is not necessarily covered by the informal notion of
robustness introduced earlier.}. There certainly are situations in which strong
expectations may override even sensory data. Such a situation can easily be
studied in everyday conversation whenever e.g. pragmatic expectations are
predominant. A similar problem occurs in experimental settings using
intentionally desynchronised video input, where lip reading information
sometimes overrides even the auditory stimulus. The problem is
witnessed as well by the difficulties usually encountered in proof-reading
one's
own text: Extremely strong expectations concerning the content usually cause
minor mistakes to be passed unnoticed.

Typically, expectations are contradictory and will be of different impact on
the progress of the analysis procedure. Hence, there is a third principle of
robust language processing upon which the human model builds. It concerns the
{\em preference-based selection} between both, competing interpretations as
well as different expectations
\cite{jackendoff:83}. Expectations have to be ranked according to their
particular strength and weighted against each other. \\

Recently linguistic research has shown a remarkable trend towards the
development of integrated models of language structure. One of the more popular
examples surely is Head-Driven Phrase Structure Grammar (HPSG
\cite{pollard:sag:94}), where syntactic and semantic descriptions are uniquely
related to each other by coreferential pointers within the framework of typed
feature structures. The strong coupling on the level of representation {\em
and} on
the level of processing (i.e. within unification) completely lacks autonomy.
The construction of a logical form
is always mediated by syntactic descriptions taken e.g. from subcategorization
information. Since syntactic and semantic restrictions are conjunctively
combined the overall vulnerability against arbitrary impairment of the input
utterances even increases: An analysis may now fail due to syntactic as well as
due to semantic reasons.

A quite similar conclusion can be drawn for construction grammar
\cite{fillmore:etal:88}, another integrated approach. It combines
syntactic, semantic and even pragmatic information in a single representation
named construction. Again, autonomy of individual description levels is missing
and even if constructions are supplied with preferential weightings derived
from
their frequency of use (as realized in SAL \cite{jurafsky:93}) robustness does
not increase.

A clearcut separation of representational levels has actually been realized in
the
cognitively motivated parser COMPERE \cite{holbrook:etal:92}
\cite{mahesh:eiselt:94}. The system aims at modelling error recovery techniques
for garden-path sentences. It uses an arbitration mechanism to decide in case
of a
conflict situation which alternative reading should be backed up. This
allows to combine early commitment decisions with the possibility to switch to
another
interpretation if necessary later on. Although the parser is guided in its
decisions by different kinds of preferences, the mapping between syntactic and
semantic representations seems to be a strict one. Accordingly, it does not
provide the necessary means for conflict resolution in all those cases of
non-standard input for which no interpretation can be established. In
particular, three different cases can be distinguished
\begin{enumerate}
  \item failure on a single level (syntax or semantics)
  \item failure on both levels (syntax and semantics)
  \item no consistent mapping between levels
\end{enumerate}
Whereas the first case might be easily accommodated by the arbitration
mechanism the latter two require the abandonment of the strict mapping and its
replacement by a preference-based module interaction.

\section{Disambiguation by Constraint Propagation}

A suitable combination of the three principles discussed above might in fact
provide the foundation for an effective use of redundancy
in parallel processing structures
\begin{itemize}
  \item {\em autonomy} guarantees a fall-back behaviour for
failures of a single module
  \item {\em expectancy-oriented analysis} facilitates the
informational exchange and
  \item {\em preference-based processing} guides the analysis towards
a promising interpretation and establishes a loose coupling between modules.
\end{itemize}
These principles, even if taken together, do not explain the almost unconscious
treatment of errors in everyday communication. To simulate a similar behaviour
a
{\em selective constraint invocation} strategy will become necessary. Then,
parsing is understood as a
disambiguation procedure, which activates only specific parts of the grammar,
if
this is deemed to be unavoidable for solving a particular disambiguation
problem.
The procedure can be terminated if a sufficiently reliable disambiguation has
been achieved even if certain conditions of the grammar have never been checked
so far. Robustness is not introduced by a {\em post mortem} retraction of
constraints
but rather by their careful invocation.

Along these lines a rudimentary kind of robustness has been achieved in the
Constraint Grammar
framework \cite{karlsson:etal:95}, a system for parsing large amounts of
unrestricted text. Constraint Grammar (CG) attempts to establish a dependency
description which is underspecified with respect to the precise identity of
modifiees. Initially, it assigns a set of morphologically justified syntactic
labels to each word form in the input sentence. Possible labels among others
are \\[-3pt]

{\tabcolsep10pt
\begin{tabular}{ll}
\verb%@+FMAINV% & the finite verb of a clause \\
\verb%@SUBJ% & a grammatical subject \\
\verb%@OBJ% & a direct object \\
\verb%@DN>% & a determiner modifying a noun to the right \\
\verb%@NN>% & a noun modifying a noun to the right \\
\end{tabular}
}
\vspace{5pt}

The initial set of labels is successively reduced by applying compatibility and
surface ordering constraints until a unique interpretation has been reached or
the set of available constraints is exhausted. In the latter case, a total
disambiguation cannot be achieved by purely syntactic means, as in the
following attachment example: \\[-3pt]

{\tabcolsep5pt
\begin{tabular}{cccccccc}
{\em Bill} & {\em saw} & {\em the} & {\em little} & {\em dog} & {\em in} & {\em
the} & {\em park} \\
\verb%@SUBJ% & \verb%@+FMAINV% & \verb%@DN>% & \verb%@AN>% & \verb%@OBJ% &
\verb%@<NOM% & \verb%@DN>% & \verb%@<P% \\
& & & & & \verb%@<ADVL% \\
\end{tabular}
}
\vspace{5pt}

In contrast to traditional grammars of the phrase structure type which
license well-formed structures according to their rule system, constraint
grammar rather happens to be an {\em eliminative approach}\footnote{With this
respect a strong
parallel between the eliminative nature of disambiguation and cohort modelling
ideas for spoken word recognition \cite{marslen-wilson:87} becomes visible. }.
Instead of imposing a normative description
on the input data it takes them as starting point and tries to find a plausible
interpretation for them.

This proceeding is motivated by the finding that language is an open-ended
system and so grammar formalisms based on a ``rigid and idealized conception
of grammatical correctness are bound to leak'' \cite[p. 37]{karlsson:95a}.
Parsing,
if understood as a disambiguation procedure, is put down to the principle of
parsimony:
The more effort is spent the better disambiguation results can be expected
and \cite[p. 39]{karlsson:95a} points to the important psycholinguistic
parallel:
\begin{quotation}
``Mental effort is needed for achieving clarity, precision and maximal
information.
Less efforts imply (retention of) unclarity and ambiguity, i.e. information
decrease. In several types of parsers, rule applications create rather than
discard ambiguities: the more processing, the less unambiguous information.''
\end{quotation}
Parsing as disambiguation can well be extended to deal with fully specified
dependency structures without loosing its promising characteristics. A complete
disambiguation of structural descriptions has
first been described for Constraint Dependency Grammar
(CDG \cite{maruyama:90a}) and simply requires to replace the monadic
categories of CG by pairs consisting of the relation name and the exactly
specified
modifiee. The mutual compatibility of modifying relations is checked against a
set of
constraints and thus the set of possible modifications is successively reduced
by a constraint propagation mechanism. Further extensions of the approach
concern the inclusion of feature descriptions, valency specifications and
valency saturation conditions \cite{harper:helzerman:94}.

Though, a closer inspection of the kind of robustness feature introduced by the
eliminative mode of operation reveals that its nature is quite accidental so
far. Which
types of deviation can be tolerated indeed, strongly depends on the rather
arbitrary sequence of constraint applications. This shortcoming seems to be
closely
connected to the fact that both formalisms lack the notion of preference so
far and therefore do not have the possibility to model the ``quality'' of a
constraint\footnote{CG at least includes heuristic constraints, which may be
activated
at a particular stage of the disambiguation procedure}. Hence, adding a
preference-based selection strategy will be one of the most pressing needs
for further improvement. Such an extension will be proposed in section 5.
Before we turn to this topic section 4 introduces a modular representation
schema along the
traditional syntax-semantics distinction. It supports the desired functional
autonomy as well as a highly interactive exchange of expectations between the
two layers.

\section{Representation Layers}

Whereas Constraint Grammar restricts itself to purely syntactic means,
an integration of simple semantic criteria into Constraint Dependency Grammar
has been proposed recently \cite{harper:etal:92}. It takes into account
sortal restrictions only, attaching them to surface syntactic relations without
aiming at modularity and autonomous behavior. In order to facilitate functional
independence it will become necessary to establish separate layers for
structural
description {\em and} constraint propagation
\begin{itemize}
  \item a {\em syntactic layer} relating word forms according to functional
surface structure
notions e.g ({\sc subject-of, dir-object-of, prep-modifier-of}, etc.) and using
constraints
on ordering, agreement, valency, and valency saturation to select among
competing
structural configuration and
  \item a {\em semantic layer} building sentence structures by means of
thematic roles
(like {\sc agent-of, instrument-of, time-of}, etc.) thereby relying upon the
argument structure of semantic predicates and their corresponding selectional
restrictions.\footnote{The proposed separation quite closely corresponds with
the one chosen
in \cite{holbrook:etal:92}}
\end{itemize}
The following small and rather rigid sample grammar illustrates the different
types of constraints needed:
\begin{enumerate}
  \item licensing conditions for modification relations \\[3pt]
\verb%sy1: cat(dep(X))=N% \\
\hspace*{2cm} {\tt $\to$ cat(synmod(X))=V $\wedge$ synlab(X)$\in$\{SUBJ,OBJ\} }
\footnote{{\tt dep(X)} refers to the modifier of a
relation, {\tt syndom(X)} and {\tt semdom(X)} to its modifiees. {\tt synlab(X)}
and {\tt semlab(X)} are the respective relation names.
{\tt cat(X), num(X), semprop(X)}, \ldots denote properties of the corresponding
node.} \\[3pt]
\hspace*{9mm}{\em A noun can modify a verb either as a subject or as a direct
object.} \\[-6pt]
  \item agreement conditions \\[3pt]
\verb%sy2: synlab(X)=SUBJ% $\to$ \verb%num(dep(X))=num(syndom(X))% \\[3pt]
\hspace*{9mm}{\em A subject agrees with its modifiee with respect to number.}
\\[-6pt]
  \item linear ordering constraints \\[3pt]
\verb%sy3: synlab(X)=SUBJ% $\to$ \verb%pos(dep(X))<pos(syndom(X))% \\[3pt]
\hspace*{9mm}{\em The subject precedes the finite verb.} \\[-6pt]
  \item compatibility constraints\footnote{In fact there is another general
compatibility
constraint implicitly built into the decision procedure and excluding ambiguous
modifying relations from being consistent: \\
\hspace*{4mm}{\tt sygen: } $\neg$ {\tt (syndom(X)=syndom(Y)} $\leftrightarrow$
{\tt dep(X)=dep(Y))} } \\[3pt]
\verb%sy4: syndom(X)=syndom(Y)% $\to$ \verb%synlab(X)% $\neq$ \verb%synlab(Y)%
\hspace*{9mm}{\em A word form cannot be modified twice by the same relation.}
\end{enumerate}
{\tt sy1} through {\tt sy3} are unary constraints, {\tt sy4} is a binary one.
Note that
constraints refer to modifying relations instead of word forms. Therefore they
are able to express admissibility
conditions on local configurations consisting of up to three nodes. Note
as well that {\tt sy3} -- even for German main clauses -- has a strong
heuristic appearance
and simply states a preference condition which additionally requires a suitable
exception handling mechanism.

In a very similar fashion semantic constraints comprise
\begin{enumerate}
  \item licensing conditions \\[3pt]
\verb%se1: cat(dep(X))=N% $\to$ \verb%cat(semdom(X))=V% $\wedge$
{\tt semlab(X)$\in$\{AG,PAT\} } \\[-6pt]
  \item selectional restrictions \\[3pt]
\verb%se2: word(semdom(X))=fressen% $\wedge$ \verb%semprop(mod(X))=animal%
\\ \hspace*{2cm} $\to$ \verb%semlab(X)=AG% \\[3pt]
\hspace*{9mm}{\em Animals do eat.} \\[6pt]
\verb%se3: word(semdom(X))=fressen% $\wedge$ \verb%semprop(mod(X))=plant%
\\ \hspace*{2cm} $\to$ \verb%semlab(X)=PAT% \\[3pt]
\hspace*{9mm}{\em Plants are to be eaten.} \\[-6pt]
  \item compatibility constraints\footnote{Again, supplemented by a general
semantic uniqueness constraint \\ \hspace*{4mm}{\tt segen: } $\neg$ {\tt
(semdom(X)=semdom(Y)} $\leftrightarrow$
{\tt dep(X)=dep(Y))}}\\[3pt]
\verb%se4: semdom(X)=semdom(Y)% $\to$ \verb%semlab(X)%$\neq$\verb%semlab(Y)%
\end{enumerate}

Adhering to the principle of autonomy both layers are designed in a way which
allows them to propagate constraints in a completely independent manner. Each
modifier is specified for two possibly different modifiees and no
cross-reference between the layers has been used so far.

In order to finally mediate the interaction between layers, a set of mapping
constraints has to be provided which sets up bidirectional correspondences

\begin{enumerate}
  \item[ ] \verb%ss1: syndom(X)=semdom(X)% $\to$ \verb%(synlab(X)=SUBJ%
\verb%semlab=AG)% \\[3pt]
\hspace*{9mm}{\em The subject of a verb is always identical to its agens.}
\\[-6pt]
  \item[ ] \verb%ss2: syndom(X)=semdom(X)% $\to$ \verb%(synlab(X)=OBJ%
\verb%semlab=PAT)% \\[3pt]
\hspace*{9mm}{\em The direct object of a verb is always identical to its
patiens.}
\end{enumerate}

It should have become obvious that the selectional restrictions as well as the
mapping constraints at best can be taken to stand for a preferential
interpretation. They surely are much to rigid to be sensibly used
within a framework of strict reasoning.

Semantic constraints need not be restricted to linguistically motivated (i.e.
universally valid) ones. In particular, domain-specific restrictions play a
crucial
role in semantic disambiguation and should urgently be incorporated whenever
possible. Here the semantic layer offers a convenient interface to a knowledge
representation component which (on demand) can contribute constraints from e.g.
specialized ontologies, referential instantiations or temporal reasoning.

\section{Weakening Constraints}

So far, one of the most striking shortcomings has been the strictly binary
nature of constraint satisfaction. Not surprisingly, it turned out to be most
inappropriate within the area of semantic modelling where hardly a constraint
can be formulated without restricting oneself to a particular, preferential
reading.

In what follows, preferences are not modelled in the usual direct
manner by emphasizing particular well-formed interpretations but rather
indirectly by putting a {\em penalty} on all remaining alternatives
which violate a constraint. For this purpose each constraint gets a
penalty factor {\tt pf} assigned reducing the confidence score in
negative cases.  Penalty factors may range from zero to one where
\begin{itemize}
  \item[ ] {\tabcolsep10pt \begin{tabular}{ll}
{\tt pf=0} & specifies a strict constraint in the classical sense and\\[3pt]
\verb%0<pf<1% & indicates a soft constraint accepting contradictory
cases \\ & with a confidence value proportional to {\tt pf} \end{tabular} }
\end{itemize}
Obviously, a value of one is meaningless because it
neutralizes the constraint. Penalty factors are combined
multiplicatively, i.e. compatibility matrices within the constraint
satisfaction problem no longer contain binary categories but {\em
confidence scores} also ranging from zero (for impossible combinations)
up to one (for combinations not even violating a single constraint).

The indirect treatment of preference by penalty factors offers a
consistent extension to the basic paradigm of constraint satisfaction.
It does not sacrifice the eliminative nature of constraints but simply
softens it. Inappropriate readings are excluded only if they violate
strict constraints. In all other cases they are downgraded to a certain
degree.

In particular, the penalty-based approach helps to tackle some
normalization problems otherwise inherently connected with the
constraint satisfaction approach: Most modifying relations (or
combinations of them) will pass a constraint simply because it is
irrelevant for that particular configuration. An
increase of goodness estimates for these cases would yield a highly
undesirable, since
unjustified reinforcement.

By assigning the penalty factors \verb%pf(sy1)=pf(sy4)=0% to the
constraints {\tt sy1} and {\tt sy4} from section 3 both are declared to
be strict ones, a fact obviously being valid for the toy-size sample
grammar which does not take into account coordinative structures. Using
\verb%pf(sy2)=0.1% the agreement condition is treated as a rather
strong one which allows exceptions only occasionally.
\verb%pf(sy3)=0.3% on the other hand results in a much more permissive
constraint justified by the fact that {\tt sy2} is meant to exclude
ungrammatical utterances but {\tt sy3} only to disfavour a marked
ordering.

On the semantic layer only the licensing constraint {\tt se1} is declared as a
strict one. The compatibility constraint {\tt se4} is weakened considerably
in order to account for double modification as in the case of anaphoric
reference. The two selectional constraints receive penalty factors of different
strength in order to model the lower probability of a plant eating something
({\tt se2}) compared to an animal being eaten ({\tt se3}): \\[3pt]
\hspace*{1cm} {\tt pf(se1)=0.0} \\
\hspace*{1cm} {\tt pf(se2)=0.1} \\
\hspace*{1cm} {\tt pf(se3)=0.7} \\
\hspace*{1cm} {\tt pf(se4)=0.5} \\[3pt]
The mapping constraints, finally, are weighted in a way which strongly favours
the {\sc subject-agens} and {\sc object-patiens} pairings nevertheless
allowing alternative interpretations e.g. in passive sentences. \\[3pt]
\hspace*{1cm} {\tt pf(ss1)=0.2} \\
\hspace*{1cm} {\tt pf(ss2)=0.3} \\[3pt]
Alternative readings can but need not be specified explicitly. In more
realistic applications though it is recommended to aim at a considerably richer
modelling, otherwise an unbalanced penalty factor as between {\tt se2} and
{\tt se3} may create a sometimes undesired strong bias by default: If both
constraints appear to be of no relevance the patience reading is clearly
prioritized. In the example chosen this corresponds to the acceptable
interpretation that an arbitrary thing is more likely to be eaten than to eat.

After having introduced penalty factors as a means of modelling preferences
constraint propagation can be extended from the classical case of strictly
binary decisions to the handling of confidence scores. The application of
penalty-weighted constraints
to a disambiguation problem now consists of two steps:
\begin{enumerate}
  \item the calculation of initial confidence scores for all combinations
of syntactic and semantic modification relations and
  \item a selection procedure pruning the search space by sorting out
unlikely interpretations
\end{enumerate}
The selection procedure is based on a local assessment function
heuristically identifying relations to be pruned. In order to not
select promising hypotheses assessment first of all should only take into
account
modification relations characterized by the following three criteria
\begin{itemize}
  \item being close to the global minimum for all modification relations,
combined with
  \item an as possible as high contrast to alternative relations and
  \item a low contrast between all the confidence scores supporting the
relation
in question.
\end{itemize}

For experimental purposes a selection procedure based on the sum of quadratic
errors
for setting scores to zero has been used. Hence, structural interpretations
violating
a high number of rather strong constraints are pruned first.

Using the toy grammar specified above together with its penalty scores
the arbitration process between syntactic and semantic evidence in simple
disambiguation problems can be studied. Thus in a sentence like \\[3pt]
\hspace*{1cm}{\em Pferde fressen Gras.} \hspace{2.3cm} {\em (Horses eat
grass.)} \hfill (2a) \\[3pt]
both layers uniformly support a single interpretation \\

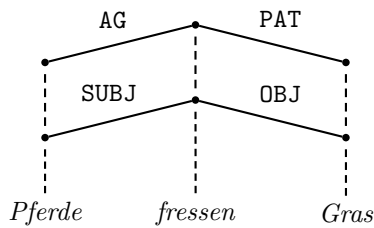
\pspicture(15,3.2)

\rput(3,0.5){\rnode{a1}{\em Pferde}}
\rput(5,0.5){\rnode{a2}{\em fressen}}
\rput(7,0.5){\rnode{a3}{\em Gras}}
\cnode*[fillcolor=black](3,1.5){0.05}{b1}
\cnode*[fillcolor=black](5,2){0.05}{b2}
\cnode*[fillcolor=black](7,1.5){0.05}{b3}
\cnode*[fillcolor=black](3,2.5){0.05}{c1}
\cnode*[fillcolor=black](5,3){0.05}{c2}
\cnode*[fillcolor=black](7,2.5){0.05}{c3}
\ncline[linestyle=dashed,nodesep=3pt,dash=3pt 2pt]{a1}{b1}
\ncline[linestyle=dashed,nodesep=3pt,dash=3pt 2pt]{a2}{b2}
\ncline[linestyle=dashed,nodesep=3pt,dash=3pt 2pt]{a3}{b3}
\ncline[linestyle=dashed,nodesep=3pt,dash=3pt 2pt]{b1}{c1}
\ncline[linestyle=dashed,nodesep=3pt,dash=3pt 2pt]{b2}{c2}
\ncline[linestyle=dashed,nodesep=3pt,dash=3pt 2pt]{b3}{c3}
\ncline{b1}{b2} \Aput{\tt SUBJ}
\ncline{b3}{b2} \Bput{\tt OBJ}
\ncline{c1}{c2} \Aput{\tt AG}
\ncline{c3}{c2} \Bput{\tt PAT}

\endpspicture
\vspace{-2mm}
Due to the strong semantic support the interpretation remains unchanged if
a marked ordering (topicalization of the direct object) is chosen (2b), an
agreement error is introduced (2c) and both deviations are combined finally
(2d). \\[3pt]
\hspace*{1cm}{\em Gras$_{PAT}$ fressen Pferde$_{AG}$.} \hfill (2b) \\
\hspace*{1cm}{\em Pferd$_{AG}$ fressen Gras$_{PAT}$.} \hfill (2c) \\
\hspace*{1cm}{\em Gras$_{PAT}$ fressen Pferd$_{AG}$.} \hfill (2d) \\[3pt]
The interpretation is retained even if its semantic support is neutralized
as in the following utterance, containing a twofold type shift. \\[3pt]
\hspace*{1cm}{\em Autos$_{AG}$ fressen Geld$_{PAT}$.} \hspace{2cm} {\em (Cars
eat money.)} \hfill (3a) \\
\hspace*{1cm}{\em Geld$_{PAT}$ fressen Autos$_{AG}$.} \hfill (3b) \\
\hspace*{1cm}{\em Auto$_{AG}$ fressen Geld$_{PAT}$.} \hfill (3c) \\[3pt]
It switches to the alternative interpretation only in the case of
combined syntactic distortions \\[3pt]
\hspace*{1cm}{\em Geld$_{AG}$ fressen Auto$_{PAT}$.} \hfill (3d) \\[3pt]
Even for the counterintuitive example \\[3pt]
\hspace*{1cm}{\em Gr\"aser$_{AG}$ fressen Pferd$_{PAT}$.} \hfill (4a) \\[3pt]
which, if desired, could be taken as a headline-style utterance,
syntactic evidence will gain the upper hand against the violation of
two selectional constraints. This interpretation, however, happens to be a
rather
fragile one and breaks immediately under arbitrary syntactic variation.

Since the selection procedure operates on a global assessment of local
structural
configurations it cannot guarantee to find an optimal and globally consistent
interpretation. The partially local mode of operation, on the other hand, can
be
expected to provide a quite natural explanation for human garden-path
phenomena.
Within the framework of preference-based disambiguation they turn out to be
a special case of contradictory situations which manifest themselves as {\em
expectation violations}: The consequences of a pruning decision
may not coincide with local confidence estimations elsewhere in the constraint
network.

Expectation violations not necessarily do indicate an erroneous situation.
They are frequently used as a speaker's intentionally chosen means to attract
the attention of the audience. This happens for instance by deviating from
an unmarked ordering to emphasize a topicalized constituent (c.f. (3b)) or
by otherwise producing unexpected utterances.

On the other hand there are the typical erroneous situations which in case
of
\begin{itemize}
  \item internal difficulties (e.g. due to early commitment strategies in
garden
path situations) might offer the possibility to initiate a reanalysis and
  \item external reasons (e.g. ill-formed input) can be used to track down the
error to find a possible remedy for it.
\end{itemize}
Note that in the latter case the situation coincides with basic observations
for the
human model: Finding an interpretation for erroneous utterances will be easier
-- in terms of effort to spent -- than detecting the error, which in
turn will be less demanding than localizing or even correcting it.

As with human language processing there will be no predefined direction for the
general flow of expectations during arbitration. Whether syntactic evidence
is propagated from the syntactic to the semantic layer or vice versa depends
only on the available
information. This seems to be in accordance with recent psycholinguistic
findings which
contest the existence of purely structural disambiguation principles
\cite{konieczny:etal:94}.

\section{Preference-based Reasoning}

Eliminating implausible interpretations by locally pruning less favoured
modification
relations represents only one, though fundamental method for the disambiguation
of natural language utterances. By selecting among modifying relations
according to negative
evidence from maximally dispreferred hypotheses, the technique fits quite well
into
the constraint satisfaction approach and achieves its robust behaviour by
avoiding extremely risky decisions on a locally
topmost reading. Taking this as a starting point the basic way of reasoning
can well be complemented by a second propagation principle based on
preference-induced
constraints. These are activated only in situations where enough positive
evidence can be derived from almost uniquely determined preferences. Since the
existence of convincing preferences in
realistic disambiguation tasks represents rather
the exception than the rule the nature of this propagation principle is
secondary.

Preference-induced constraints consist of implications $P \to_p C$ which,
given enough evidence for the unary precondition $P$, require the possibly
binary
constraint $C$ to hold. Constraints of this type can be used to model e.g. the
higher-order conditions which in many mapping situations involve more than two
relations. \\[3pt]
{\tt pss1: word(syndom(X))=im} $\wedge$ {\tt synlab(X)=PHEAD} \\ \hspace*{2cm}
$\wedge$
{\tt semprop(dep(X))$\in$\{TEMP,LOC\} }\\
\hspace*{11mm}$\to_p$ {\tt synlab(Y)=PMOD} $\wedge$ {\tt dep(Y)=syndom(X)} \\
\hspace*{2cm} $\wedge$ {\tt semdom(Z)=sydom(X)} $\to$ {\tt semlab(Z)=PART-OF}
\\[3pt]
\hspace*{11mm}{\em A prepositional phrase headed by the word form ``im'' fills
a semantic \\ \hspace*{11mm}{\sc part-of} slot} \\[3pt]
In postnominal positions like \\[3pt]
\hspace*{1cm}{\em Dann nehmen wir die erste Woche im Mai.}  \hfill (5) \\
\hspace*{25mm}{\em (Let's take the first week in may.)}\\[3pt]
this constraint puts a preference on the lower attachment since a part-of
relation
usually is not licensed as an argument position for verbs.

Preference-induced constraints can also be used to modify value assignments at
certain nodes in the constraint network without the necessity to copy them. By
applying
this technique, phrasal feature projections can be modelled in order to build
up descriptions for partial dependency trees \\[3pt]
{\tt pss2: synlab(X)=DET} \\ \hspace*{11mm}$\to_p$ {\tt
case(syndom(X)):=case(syndom(X))}
$\cap$ {\tt case(dep(X))} \\[3pt]
\hspace*{11mm} {\em A noun group carries the intersection of possible case
features \\ \hspace*{11mm} found at its members.} \\[3pt]
Preference-induced constraints introduce a kind of inhibitory mechanism
to the disambiguation procedure: Already preferred interpretations is
given the chance to propagate their consequences over the network thus
possibly leading to further suppression of alternative readings.

\section{Conclusion}

Combining the eliminative nature of a disambiguation procedure with a system
architecture supporting bidirectional arbitration between syntactic and
semantic evidence has turned out to be a key factor for achieving a higher
level of robustness in  language understanding. While the disambiguation
paradigm provided the basic fall-back behaviour (an arbitrary utterance will
get a description assigned) and the possibility to prune the search space
towards a least disfavoured reading, the parallel arrangement of modules allows
to interactively exchange expectations and thus
bypassing local interpretation difficulties. By modelling a preference
distribution based on penalty factors, the desired robust behaviour can be
demonstrated at least for very simple sample utterances. Although no conclusive
judgement about the feasibility of the approach can be given until the
experimental setting has been scaled up to a fairly realistic problem size, a
remarkable qualitative advance over comparable approaches becomes evident even
on this elementary level:
\begin{itemize}
  \item The approach departs from a predefined sequential arrangement of
modules in favour of a strictly symmetrical architecture consisting of
autonomous components for syntax and semantics.
  \item It allows to treat syntactic ill-formedness and semantic deviations by
providing a mechanism for mutual compensation. Syntactically anomalous
utterances can be understood as long as there is enough semantic and/or
pragmatic evidence. In order to communicate novel or unusual content a
sufficiently high degree of syntactic support is required.
  \item Insufficient modelling information on any one of the processing layers
might well result in the selection of an odd interpretation but will not cause
the language processing unit to break down entirely.
  \item Robustness is not an add-on feature of an otherwise temperamental
procedure but falls out from the basic properties of the processing mechanism.
\end{itemize}
Since structural disambiguation by constraint
satisfaction likewise lends itself to the creation of time sensitive
parsing procedures \cite{menzel:94a}, in the long run it might provide a
unifying foundation to build language processing systems upon which
embody aspects of robustness against such different disruptive factors
as syntactically ill-formed input, metaphorical use and dynamic time
constraints.

\bibliographystyle{abbrv}

\end{document}